# Leveraging Auxiliary Text for Deep Recognition of Unseen Visual Relationships

**Gal S.Kenigsfield**  SGALK87@CAMPUS.TECHNION.AC.IL
*Department of Electrical Engineering*
*Technion*
*Haifa, Israel*

**Ran El-Yaniv**  RANI@CS.TECHNION.AC.IL
*Deparment of Computer Science*
*Technion*
*Haifa, Israel*

## Abstract

One of the most difficult tasks in *scene understanding* is recognizing interactions between objects in an image. This task is often called *visual relationship detection* (VRD). We consider the question of whether, given auxiliary textual data in addition to the standard visual data used for training VRD models, VRD performance can be improved. We present a new deep model that can leverage additional textual data. Our model relies on a shared text–image representation of subject-verb-object relationships appearing in the text, and object interactions in images. Our method is the first to enable recognition of visual relationships missing in the visual training data and appearing only in the auxiliary text. We test our approach on two different text sources: text originating in images and text originating in books. We test and validate our approach using two large-scale recognition tasks: VRD and Scene Graph Generation. We show a surprising result: Our approach works better with text originating in books, and outperforms the text originating in images on the task of unseen relationship recognition. It is comparable to the model which utilizes text originating in images on the task of seen relationship recognition.

## 1. Introduction

Scene Graph Generation (SGG) is the task of inferring a *scene graph* (SG) given an image. An SG is a topological structure of a scene where the nodes represent the objects and the edges represent the relationships between pairs of objects. Inferring an SG allows the extraction of information from the image (e.g., regional descriptions, global descriptions, labels etc.). For example, in Figure 1 we see the SG of an image containing three relationships among four objects. The SGG task, which incorporates computer vision and natural language understanding, belongs to a family of tasks that require abstract capabilities that are deeper and much more challenging than standard image classification or tracking/detection tasks (Johnson et al., 2015; Krishna et al., 2016).

A straightforward approach to generate an SG is to decompose the task into subtasks such that the SG is assembled from a set of inferred relationships between all object pairs in the image. The subtask of inferring the interaction or relationship between a pair of given objects is called *visual relationship detection* (VRD). For example, given the objects `player` and `ball` in Figure 1, a correct model should infer the relationship `dribbling`.

We introduce RONNIE (= **r**ecognition **o**f u**n**seen **v**isual **r**elationships), a model designed to utilize auxiliary text and enable recognition of unseen visual relationships. The first step in our approach is to pre-processes text describing images or visual scenes into a subject-predicate-object database, and from the parsed text we assemble an object-relationship mapping. We define object-relationship mapping as a function from a pair of objects to a set of predicates $\Phi(o_i, o_j) = \{r_{i \to j}^k\}_{k=1}^K$, where $o_i$ and $o_j$ are objects and $r_{i \to j}$ is their relationship. . We compute embeddings for all predicates and use an attention mechanism to leverage the embeddings for relationship detection. By utilizing the parsed text, our model is able to recognize relationships even if they appear less frequently completely absent from the training data. We demonstrate how our approach facilitates recognition of unseen visual relationships on the Visual Genome (VG) dataset (Krishna et al., 2016). We train our model on VG-200 – a filtered version of VG containing 150 objects and 50 relationships. VG-200 was introduced by Xu et al. (2017). We evaluate RONNIE on a new variant of VG called VG-Min100, which we introduce in Section 5 to evaluate the task on unseen classes.

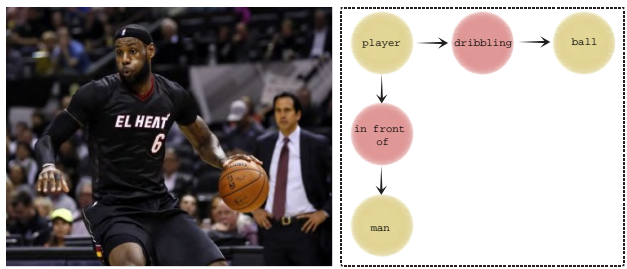

Figure 1: Left: an image from VG with ground truth bounding boxes and labels. Right: an example of a scene graph. The graph contains the objects `player` and `ball`. The interaction between them, which we would like to infer is `dribbling`, is absent from the training data.

## 2. Related Work

### 2.1 Scene Graph Generation

Scene graphs (SGs) were first introduced by Johnson et al. (2015), who utilized them for image retrieval. An SG is a topological description of a scene with the nodes corresponding to objects and the (directed) edges corresponding to the relationship between pairs of objects. An early approach proposed by Lu et al. (2016) detected all the objects in the scene and later utilized object appearances to detect relationships between objects. Xu et al. (2017) used graph-based inference to propagate information in both directions between objects and relationships. Zellers et al. (2018) investigated recurring structures in VG-SG and employed a global context network to predict the graphs. They also introduced a strong frequency baseline based on VG statistics. Herzig et al. (2018) proposed a permutation-invariant prediction model.

### 2.2 Visual Relationship Detection

Early studies in visual relationship detection tended to rely on data statistics (Mensink et al., 2014), or adopt a joint model for a relationship triplet (subject-relationship-object). Lu et al. (2016) showed how to utilize a relationship embedding space from the subject and object appearance model for visual relationship prediction. Zhuang et al. (2017) and Zhang et al. (2017) used visual embedding networks, which embed objects in a low-dimensional space and integrate them as context for VRD.

To solve the SGG task, we follow Lu et al. (2016) and decompose it into two independent subtasks: (1) Detecting the set of $N$ objects, and (2) detecting $O(N^2)$ relationships (potentially, between all object pairs). Due to the complexity of VRD, integrating data from various sources may be necessary. All recent approaches demonstrated success using small vocabularies, e.g., 150 objects and 50 relationships. We introduce a novel approach to integrating data from various sources and enable scaling VRD onto larger vocabularies.

#### 2.2.1 LARGE-SCALE VISUAL RELATIONSHIP DETECTION

Real-world visual scenes are populated with a vast number of objects and visual relationships. Systems designed to recognize visual relationships are usually limited to a fixed number of pre-defined classes. This limitation is in part due to the difficulty in acquiring training data as well as sparsity along the long tail of the object-relationship distribution.

Zhang et al. (2018) were the first to demonstrate large-scale visual relationship detection by constructing a model for the challenging VG-80K dataset (see Section 5). Our work is inspired by the seminal paper by Frome et al. (2013) who showed how to predict the labels of visual objects that were not present in the training set. This was accomplished by utilizing auxiliary word embedding. Our model can be viewed as a substantial extension of the Frome et al. (2013) result for *relationships* (rather than *objects*).

## 3. Problem Formulation

Following Zellers et al. (2018) and Chen et al. (2019), we define an SG for a given image $I$ as a directed graph $G_I \triangleq (O, R, B)$, where $O \triangleq \{o_1, o_2, \ldots, o_n\}$ is a set of (visual) objects appearing in $I$, $R \triangleq \{r_{1\to 2}, r_{1\to 3}, \ldots, r_{(n-1)\to n}\}$ is a set of directed edges representing (non-symmetric) relationships, potentially between all object pairs, and $B \triangleq \{b_1, b_2, \ldots, b_n\}$ is a set of bounding boxes, where $b_i \triangleq (x, y, w, h)$ is the bounding box of object $o_i$. The bounding box definition is standard, with $(x, y)$ being the center coordinates of the box, and $w, h$ its width and height, respectively. For example, Figure 1 (right) is the SG of the image in Figure 1 (left).

Setting $p(G|I) \triangleq p(B, O, R|I)$, we decompose the probability distribution $p(G_I|I)$ of the graph $G_I$ into three components:

$$p(G_I|I) = p(B|I)p(O|B, I)p(R|O, B, I). \tag{1}$$

This decomposition, and the three components, motivate three computation steps that are sufficient for assembling the SG. The first term,

$$p(B|I) = \Pi_{i=1}^{N} p(b_i|I), \tag{2}$$

corresponds to the first step whereby the bounding boxes in the image are identified. Given these bounding boxes, the second term,

$$p(O|B, I) = \Pi_{i,j=1}^{N} p(o_i|b_i), \tag{3}$$

corresponds to predicting class labels for the objects (within their bounding boxes). The third term,

$$p(R|O, B, I) = \Pi_{i=1}^{N} p(r_{i\to j}|o_i, o_j), \tag{4}$$

corresponds to the last step where relationships are predicted for object pairs. This task of identifying a relationship given two objects is called *visual relationship detection* (VRD) (Lu et al., 2016).

Following (Lu et al., 2016; Xu et al., 2017; Zellers et al., 2018; Chen et al., 2019), we consider a supervised structure learning approach to generating SGs. Given a set of training examples, $S_m \triangleq \{(I^{(i)}, SG^{(i)}), i = 1 \ldots m\}$, where $I^{(i)}$ is an image and $SG^{(i)}$ is its corresponding SG, the goal is to train a model to predict SGs for unseen images. The common performance measure for both the VRD and SGG tasks is *recall at K* (R@K) (Lu et al., 2016), which computes the fraction of correctly predicted object-relationship-object triplets among the top-K confident predictions.

**Recognition of unseen relationships** We now formulate the task of recognizing unseen visual relationships, and then describe the evaluation metrics adopted for this task. Given a set of training examples $\mathcal{S}_m$, containing a set of relationships $R_{train}$, and an auxiliary text corpus $T$ containing a set of relationships $R_T$, we follow the same training protocol as for SGG; the only difference is that here we use $T$ to facilitate detection of relationships unseen in $R_{train}$. To test our model's effectiveness on the task of recognizing unseen visual relationships, we propose a new variant of VG, called *VG-Min100*, which is based on a relationship set $R_{Min100}$ containing relationships unseen in $R_{train}$. At test time we aim to assign the correct relationship $r^*$ to a pair of objects even though $r^*$ is not necessarily contained in $R_{train}$. We use top-5 accuracy, top-10 accuracy, and R@K as evaluation metrics for our unseen visual relationship predictions.

## 4. RONNIE

Our method, acronymed RONNIE (= **r**ecognition **o**f u**n**see**n** v**i**sual r**e**lationships), is schematically illustrated in Figure 2. It comprises four main components: (a) an object detector, (b) an object attention mechanism, (c) an object-relationship mapping (ORM), and (d) a relationship attention mechanism.

For the object detector, one can use any known detector such as Faster-RCNN (Ren et al., 2015), YOLO (Redmon et al., 2016) or RetinaNet (Lin et al., 2017). We now describe all other (novel) components (we used Faster-RCNN, see below).

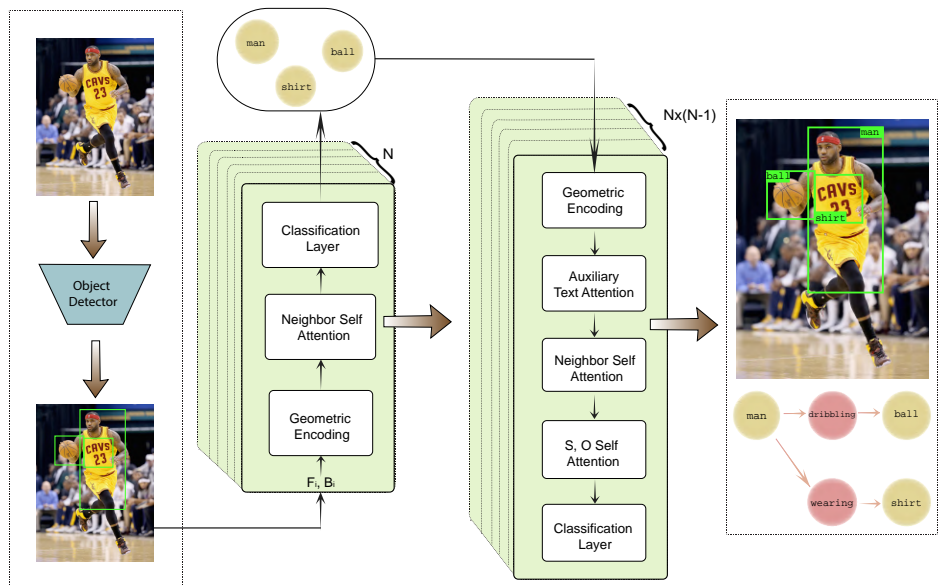

Figure 2: An overview of RONNIE. Our model comprises four components: (1) an object detector, (2) an object attention mechanism, i.e., neighbor self-attention mechanism, (3) an object-relationship mapping (ORM), and (4) a relationship attention mechanism. An image is fed into an object detector to extract visual features, the visual features of the objects go through a self-attention mechanism and then are classified. At this stage, we pool visual features for the relationships using a CNN. We extract relationship candidates given the object classes we detected with the ORM. Lastly, we feed the top-K relationship candidates into a relationship attention mechanism to leverage the linguistic text and classify the relationships.

### 4.1 Attention Mechanism

Our model utilizes attention mechanisms for various purposes. We now describe a general attention procedure (Schuster et al., 2015). Let $q \in \mathbb{R}^n$ be a query sample and, $C = [c_1, \ldots, c_k]$, $C \in \mathbb{R}^{k \times n}$, be the context samples. First, we obtain the attention coefficient $a_i$: $a_i = C \cdot q$, $a_i \in \mathbb{R}$. Next, we calculate the attention weights $w_i$: $w_i = \text{Softmax}(a_i)$, $w_i \in \mathbb{R}$. The attention vector $v = \frac{1}{k} \sum_{i=1}^{k} w_i c_i$ is the weighted sum of the context vector and the attention weights. The final vector $q'$ is a concatenation of the query and attention vector that we fuse using a linear layer:

$$q' = W_{att} \cdot [q, v], q' \in \mathbb{R}^n. \tag{5}$$

Throughout the paper we denote this attention procedure as $q' = A(q, c)$.

### 4.2 Detecting Objects

We now describe our object detection methodology. Given an image, the output of the object detector is a set of object proposals, as well as their corresponding bounding boxes, $B = \{b_1, b_2, \ldots, b_n\}$, with matching visual features, $F \triangleq \{f_1, \ldots, f_n\}$, extracted from the detector, such that $f_i \in \mathbb{R}^d$ (so $F^0 \in \mathbb{R}^{n \times d}$). To enrich $F$, we utilize two elements: (1) spatial information, and (2) a self-attention mechanism over neighboring visual features, i.e., neighbor self-attention. To combine spatial information with the visual features, we project each bounding box quadruplet $(x, y, w, h)$ (recall that $x$ and $y$ are the center coordinates of the bounding box, $w$ is the width, and $h$ is the height), using a single linear layer $f_B(B) = B \cdot W_{spat} + b_{spat}$, where, $B \in \mathbb{R}^{n \times 4}$ and $W_{spat} \in \mathbb{R}^{4 \times r}$. The enriched spatial and visual feature vector, $F'$, is obtained by concatenation, $F' \triangleq [F, f_B(B)]^1$, so $F' \in \mathbb{R}^{n \times (d+r)}$. To further enhance $F$, we employ a self-attention mechanism (Vaswani et al., 2017) over the visual features. Each $\{f_i^1\}_{i=1}^n \in F^1$ interacts with its neighboring visual features, $\{f_1^1, f_2^1, \ldots, f_n^1\}_{k=1, k \neq i}^n$, as follows.

$$f'_i = A_{obj,self}(f_i, [f_1, \ldots, f_{j,n}]). \tag{6}$$

We apply this mechanism to all bounding boxes such that $F' = [f'_1, \ldots, f'_n]$. Our final object prediction is $\hat{O} = \text{Softmax}(W_o \cdot F' + b_o)$, where $W_o \in \mathbb{R}^{n \times |\mathcal{O}|}$.

### 4.3 Object-Relationships Mapping

Our model relies on utilizing auxiliary text. We consider two different sources of auxiliary text: (1) text describing images from various captioned image datasets (Krishna et al., 2016; Lin et al., 2014; Plummer et al., 2015), denoted as RONNIE , and (2) text from books taken from Project Gutenberg (Lebert, 2008) denoted as RONNIE$_{gut}$ . We follow Schuster et al. (2015), and parse the text into subject-relationship-object (s-r-o) triplets. For example, in Figure 1, `player` is the subject, `dribbling` is the relationship, and `ball` is the object. Denoting the subject-relationship-object distribution in the auxiliary text by $P$, we can express and estimate the empirical subject-relationship-object distribution, $\hat{P}$, using standard counting statistics,

$$\hat{P}(r_{s \to o}|s, o) \triangleq \frac{Count(s - r - o)}{Count(s - o)}. \tag{7}$$

We define an object-relationship mapping that maps object pairs to a set of relationships and their corresponding probabilities,

$$\Phi \colon S \times O \to \{(\mathcal{R}_{S \to O}, \hat{P}(R_{S \to O}|S, O))\}, \tag{8}$$

where $S$ and $O$ are the set of object categories, $\mathcal{R}_{S \to O}$ is the set of relationships, and $\hat{P}(R_{S \to O}|S, O)\}$ is their corresponding probabilities; for example, consider an object pair from Figure 3(a): $\Phi(man, helmet) = \{$'wearing': 0.54,..., 'stands with':1e-4$\}$. To utilize the output of $\Phi$, we use a pre-trained word embedding model introduced by Mikolov et al. (2013), such that the output of the object-relationship mapping, $\hat{R}_{S \to O}$, is fed into an embedding layer, namely, a relational embedding layer.

### 4.4 How to Recognize Relationships

To facilitate recognition of visual relationships we combine three elements that enrich the visual features: (1) geometric encoding, to better express geometric relationships, (2) an object-relationship mapping that maps object pairs to a set of relationship candidates extracted from the auxiliary text, and (3) a subject-object attention mechanism, which utilizes the objects' visual features. First, we extract visual features from the union of all corresponding bounding boxes $\{b_i \cup b_j\}_{i,j=1, i \neq j}^{N \times (N-1)}$. We pool the visual features, $f_{i \to j}$, by applying an ROI-Align function, a two-layer convolutional neural network (CNN) and a single linear layer, $f_{i \to j} \in \mathbb{R}^d$. We denote the set of visual features by $F_{i \to j} \triangleq [f_{1 \to 2}^0, f_{1 \to 3}^0, \ldots, f_{(n-1) \to n}^0]$, where $F_{i \to j} \in \mathbb{R}^{m \times d}$.

#### 4.4.1 GEOMETRIC ENCODING

Many relationships appearing in datasets such as VG-200 are geometric (Zellers et al., 2018), e.g., `under`, `on top of`, `next to`. Following Hu et al. (2018), to support recognition of geometric relationships, we utilize a geometric relationship encoding, $g_{i \to j} = [\frac{x_i - x_j}{w_i}, \frac{y_i - y_j}{h_i}, \frac{w_j}{w_i}, \frac{h_j}{h_i}] \cdot W_{geo} + b_{geo}$, where $W_{geo} \in \mathbb{R}^{4 \times r}$ is a learned weight matrix, $v_g \in \mathbb{R}^r$, and $(x_i, y_i, w_i, h_i), (x_j, y_j, w_j, h_j)$ are the subject and object bounding boxes, respectively. We denote the set of geometric encoding features $g_{i \to j} \in \mathbb{R}^r$, and apply concatenation of these features such that, $f'_{i \to j} = [f_{i \to j}, g_{i \to j}]$, where $F'_{i \to j} \in \mathbb{R}^{m \times (d+r)}$.

---

1. All concatenation operations are column-wise concatenation unless stated otherwise.

### 4.4.2 AUXILIARY TEXT UTILIZATION

We now describe how to utilize the auxiliary text, which is a key element of this work. The intuition is that fusion of the auxiliary text into the SGG training process facilitates recognition of unseen relationships. We use object relationship mapping ($\Phi$) to combine information from the text with the visual information. Consider Figure 3(a). The set of object detections in the given image is $\hat{O}$, where $o_i$ signifies man, and $o_j$ signifies helmet. We map all possible object pairs $\{o_i, o_j\}_{i,j=1}^N$ using $\Phi$. We obtain $\hat{R}_{i \to j} = \Phi(o_i, o_j) = \{\hat{r}_{ij}^k, \hat{p}_{ij}^k\}$ – that is, the set of relationships and their probabilities. Using $\{\hat{p}_{i \to j}^k\}_{k=1}^K$, we rank the relationship candidates in descending order, and randomly draw a subset of $k$ relationships from the top-M relationships to prevent our model from stagnating. Next we feed the relationship candidates to a relational embedding layer. The output of the relational embedding layer is a distributed word representation, $\{v_{r_{i \to j}}^k\}_{k=1}^K$, where $v_{r_{i \to j}}^k \in \mathbb{R}^e$. The relational embeddings are then projected into the dimension of $F'_{i \to j}$ by using a single linear layer, $V_{r_{i \to j}} = [v_{r_{i \to j}}^0, v_{r_{i \to j}}^1, \ldots, v_{r_{i \to j}}^k] \cdot W_{txt} + b_{txt}$, where $W_{txt} \in \mathbb{R}^{e \times (d+r)}$, and $V_{r_{i \to j}} \in R^{k \times (d+r)}$. To leverage the relational embeddings with the visual features, we apply an attention mechanism $f''_{i \to j} = A_{rel,aux}(f'_{i \to j}, V_{r_{i \to j}})$.

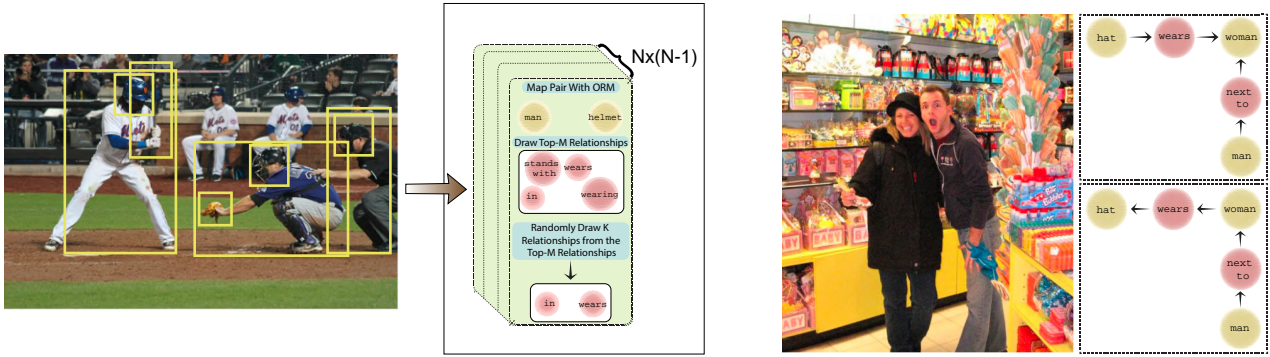

(a) Left: an image from VG with its ground truth bounding boxes. Right: for each pair of objects we perform the procedure illustrated on the right to extract relationships from the auxiliary text. We perform three steps: (1) feed the subject-object pair to $\phi(\cdot)$, (2) rank the top M relationships from $\phi$, and (3) randomly draw K relationships from the top-M relationships.

(b) Left: an image from VG. Right: an SG obtained by RONNIE without a subject-object attention mechanism; the triplet {hat, wearing, woman} in bold is a obviously the result of subject-object confusion. On the bottom right: an SG obtained by RONNIE with a subject-object attention mechanism that was able to overcome this confusion.

Figure 3

### 4.5 Subject-Object Confusion

In the process of relationship detection we propose the union of possible pairs of bounding boxes as ROIs for the visual relationship. One problem that arises from this procedure is subject-object confusion. In Figure 3(b), the left graph is obtained by RONNIE without a subject-object attention mechanism and we can see that there is confusion between the subject woman and object hat. To resolve this problem we propose our subject-object attention mechanism. For each union of bounding boxes, we apply an attention mechanism such that $f'''_{i \to j} = A_{rel,s-o}(f''_{i \to j}, [[f'_i, f_j], [f_j, f_i]])$. Our final relationships prediction is $\hat{R} = \text{Softmax}(W_r \cdot F'''_{i \to j} + b_r)$. We also train an embedding layer for unseen relationship recognition (see Section 5), such that $\hat{V}_r = W_{r,e} \cdot F'''_{i \to j} + b_{r,e}$

### 4.6 Loss Function

Our loss function is designed to optimize both SG generation and recognition of unseen visual relationships by leveraging the availability of auxiliary text. The main idea is to encourage our model to learn visual relationship representations that are similar to text-induced representations. Such representations enable inferring relationships whose labels are absent in the visual data but yet are still similar semantically to relationships that appear in the auxiliary text. Our loss function contains several components and relies on both the cosine and the cross-entropy (CE) loss functions. The use of the cosine loss for linking vision and text embeddings was found to be very effective by Shalev et al. (2018). The cosine loss function can be defined in terms of the standard *cosine similarity* metric, which for two vectors, $\mathbf{u}$ and $\mathbf{v}$, is $\cos(\mathbf{u}, \mathbf{v}) \triangleq \frac{\mathbf{u} \cdot \mathbf{v}}{||\mathbf{u}||||\mathbf{v}||}$.

The cosine loss function is $L_{\cos}(\mathbf{u}, \hat{\mathbf{u}}) \triangleq 1 - \cos(\mathbf{u}, \hat{\mathbf{u}})$. Thus, two labels are considered semantically similar if and only if their corresponding embeddings are close, namely, $L_{\cos}$ is small. The second element used to define our loss function components is the standard CE loss $L_{CE}(\mathbf{u}, \hat{\mathbf{u}}) \triangleq -\mathbf{u} \log(\hat{\mathbf{u}})$. The proposed loss function is,

$$L(O, R, V_R, \hat{O}, \hat{R}, \hat{V}_R) = \lambda_1 L_{CE}(O, \hat{O}) + \lambda_2 L_{CE}(R, \hat{R}) + \lambda_3 L_{\cos}(V_R, \hat{V}_R), \qquad (9)$$

where $\hat{O}$, $\hat{R}$ and $\hat{V}_R$ are the object predictions, relationship predictions, and relationship representations, respectively. The first and second terms are standard when considering multiclass classification tasks (in our case, both object and relationship predictions). The third term penalizes representations ($\hat{V}_R$) that diverge from the corresponding ground truth word embedding ($V_R$).

## 5. Experimental Settings

We present various experiments and comparisons of RONNIE to several baselines. In addition to the full RONNIE model as described in Section 4, we consider a scaled-down version of this model, referred to as RONNIE$_0$, which does not utilize the auxiliary text. We experiment with two subsets of VG-80K. The first is the subset introduced by Xu et al. (2017) and called *VG-200*. The second is a new VG-80K subset that we term *VG-Min100* (see details below).

### 5.1 Datasets

- **VG-200.** Introduced by Xu et al. (2017), this set is a filtered version of VG-80K containing the most frequent 150 objects and most frequent 50 relationships. We train both RONNIE and RONNIE$_0$ on VG-200 and compare them to four recently proposed models (Zellers et al., 2018; Xu et al., 2017; Lu et al., 2016; Zhang et al., 2018).

- **VG-Min100.** We consider a new filtered version of VG80K, containing all objects and relationships occurring at least 100 times in VG-80K. Following Zhang et al. (2018), we also remove non-alphabetical characters and stop words, resulting in 1,845 objects and 450 relationships. We follow Johnson et al. (2015) and split the data into 103,077 training images and 5,000 testing images. The VG-Min100 contains 400 relationships unseen during training, thus it allows us to demonstrate how RONNIE facilitates the recognition of visual relationships that were not included in the training set.

We train our model on VG-200 and evaluate it on VG-200. To demonstrate RONNIE's ability to recognize unseen visual relationship classes, we also evaluate RONNIE on VG-Min100.

### 5.2 Recognition of Unseen Relationships

To evaluate recognition of unseen relationships, we use VG-Min100. To compare our model to previous approaches (Xu et al., 2017; Zellers et al., 2018), we re-train the earlier models with a relational embedding layer and the loss function presented in Section 4. As mentioned earlier, we evaluate performance using R@K, and top-5 and top-10 accuracies. We focus on predicting visual relationships; thus, at test time we provide ground truth boxes and object labels. To predict the relationships we utilize cosine similarity (see above). Two vectors are considered close if their cosine similarity is close to one. For each vector, the model produces a relationship representation, $v_r$, and we compute $\hat{r} \triangleq \text{Softmax}(\cos(\hat{V}_r, V_R^{GT}))$, where $V_R^{GT} \in \mathbb{R}^{e \times |R|}$ are the word embeddings representing the relationship categories in our test set. We adopt the definition of Zhang et al. (2018) for the long-tail distribution of visual relationships in VG, namely, rare relationships, with less than 1,024 instances. We follow the same split of VG-80K as described in Zhang et al. (2018) regarding VG-Min100.

### 5.3 Scene Graph Generation

For VG-200, we use the same evaluation protocol used by Zellers et al. (2018) and Xu et al. (2017) who considered the following two tasks. In the first task, denoted **Pred-Cls**, the goal is to predict relationship labels, given the correct labels for subjects and objects. The objective in the second and harder task, denoted **SG-Cls**, is to predict subject and object labels, given their correct bounding boxes as input. In addition, correct relationships must be predicted.

## 6. Results

### 6.1 Recognition of Unseen Relationships

The task of recognizing unseen relationships is the main focus of this paper. Our results for RONNIE and the baselines appear in Table 1. The table has two sections for all relationships (both seen and unseen) from the VG-Min100 dataset, and for rare (long-tail) unseen relationships. The table presents R@K, top-5 and top-10 accuracy results for RONNIE and for all baselines. It is evident that RONNIE has no competition at all in both types of relationships. Moreover, in the case of rare relationships, none of the baselines including $RONNIE_0$ yields meaningful results despite the enhancements we applied to them (see Section 4). Lastly, consider the two last rows in Table 1 as a comparison between the two text sources. This result is particularly interesting as it shows that text originating in prose (books) has clear advantage (using our model) over text originating in image captions.

| Model | Top-5 | Top-10 | R@50 | R@100 |
|---|---|---|---|---|
| | All Classes | | | |
| $Motifs$ | 68.5 | 74.3 | 52 | 55 |
| $RONNIE_0$ | 74.5 | 79.8 | 57 | 62 |
| RONNIE | 78 | **82.4** | **65** | **69** |
| $RONNIE_{gut}$ | **78.25** | 81.87 | 62 | 66.5 |
| | Rare Relationships | | | |
| | Top-5 | | | |
| $Motifs$ | 22 | | | |
| $RONNIE_0$ | 30 | | | |
| RONNIE | 37.9 | | | |
| $RONNIE_{gut}$ | **44.4** | | | |

Table 1: SGG results on VG-200.

| | SG-Cls | | Pred-Cls | |
|---|---|---|---|---|
| Model | R@50 | R@100 | R@50 | R@100 |
| LP Lu et al. (2016) | 11.8 | 14.1 | 27.9 | 35 |
| IM Xu et al. (2017) | 21.7 | 24.4 | 44.8 | 53 |
| Motifs Zellers et al. (2018) | 35.8 | 36.5 | 65.2 | 67.1 |
| LS Zhang et al. (2018) | 36.7 | 36.7 | **68.4** | 68.7 |
| $RONNIE_0$ | 36.2 | 37 | 67.9 | 68 |
| **RONNIE** | **37** | **37.7** | 68 | **69.1** |
| $RONNIE_{gut}$ | 36.4 | 37.5 | 68 | 68.9 |

Table 2: Results on VG-Min100. All classes refer to all visual relationships with $\# occurrences > 1,024$. The long-tail classes follow Zhang et al. (2018)'s definition $\# occurrences < 1,024$.

### 6.2 Scene Graph Generation

In Table 2, we present results achieved by RONNIE on VG-200. We compare its performance to four previous methods (Lu et al., 2016; Xu et al., 2017; Zellers et al., 2018; Zhang et al., 2018), all of which do not utilize auxiliary text (see Section 2). To examine the effect of utilizing the auxiliary text with object-relationship mapping, consider the second-to-last row in Table 1. On the SG-Cls task, RONNIE achieved an R@100 of 37.7, and an R@50 of 37, outperforming all baselines. In general, it is evident that RONNIE is slightly better than all baseline methods at SGG. We emphasize, however, that RONNIE is using auxiliary text that cannot be utilized by the other contenders.

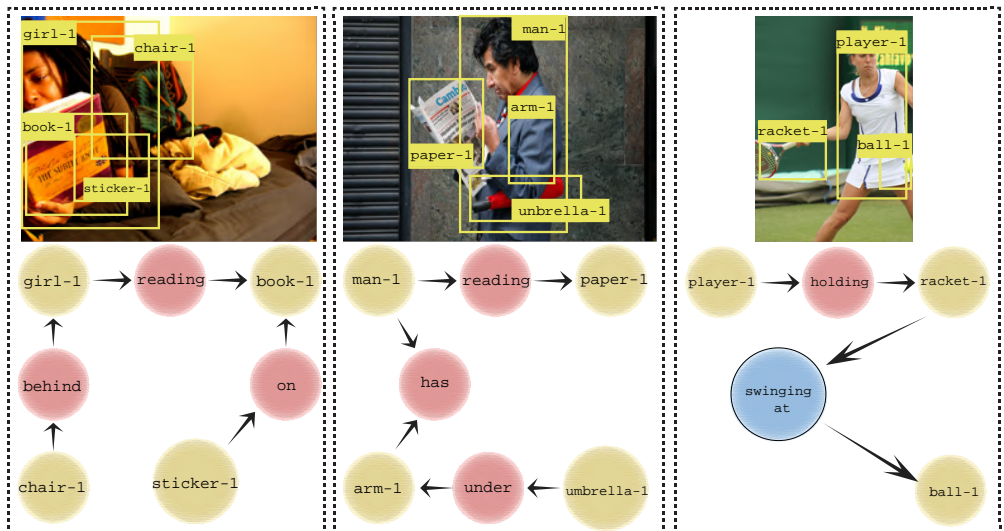

Figure 4: Qualitative results from our RONNIE model. The blue circle is unseen relationship class recognized by RONNIE and the red circles are seen classes.

| Mechanism | | | | R@50 | R@100 |
|---|---|---|---|---|---|
| Object attention | Geometric encoding | Relationship geometric encoding | Subject-object attention | | |
| ✓ | | | | 35.5 | 36.3 |
| ✓ | ✓ | | | 35.6 | 36.3 |
| ✓ | ✓ | ✓ | | 36 | 36.7 |
| ✓ | ✓ | ✓ | ✓ | 36.2 | 37 |

Table 3: Results of the SG-CLs setup on VG-200. For each row we add a mechanism, as described in Section 4, to our baseline model.

| Relationship | Top-5 acc | # synonyms | # synonyms instances |
|---|---|---|---|
| driving on | 96 | 2 | 2,004 |
| parked in | 77.6 | 12 | 6,556 |
| growing in | 57 | 6 | 3,664 |
| riding on | 46.6 | 3 | 525 |
| hold up | 27.3 | 6 | 45,580 |
| standing by | 7.3 | 14 | 20,630 |
| connected | 25 | 1 | 10,096 |
| cutting | 50 | 0 | 874 |
| standing behind | 25.8 | 14 | 20,603 |
| walking | 22.3 | 14 | 9,343 |

Table 4: Results of the top 10 relationships in the long-tail distribution. The metric we use is top-5 accuracy.

#### 6.2.1 EXAMINING SPECIFIC RARE RELATIONSHIPS

In this section we focus on specific rare relationships. Of the set of rare relationships whose occurrence count in the entire dataset was less than 1,024, we inspect the top 10. Consider Table 3 showing the top-5 test accuracy in predicting each rare relationships. For example, row 11 corresponds to the `standing by` relationship, which has 14 near-synonyms in VG-Min100 (e.g., `standing on`, `standing on top of`, `standing`, `standing next to`). Our method was able to recognize this relationship with a low 7.3% top-5 accuracy. This poor performance is perhaps not surprising given the relatively large number of near-synonyms, and the number of times they appear in the dataset (see the last column of Table 2). Moreover, one of these synonyms, namely, `standing on`, appears extremely frequently in the VG-Min100 dataset (14, 185 instances). In contrast, the relationship `driving on`, on which RONNIE achieved a top-5 accuracy of 96%, has only two other near-synonyms (`driving` and `driving down`) that are also rare. While these extreme cases may hint that there is some regularity in the functional relationship between accuracy achieved and the number of synonyms and their occurrence rate, we note that there exist severe failures even in cases where there are few rare synonyms, or no synonyms at all (see, e.g., `cutting`).

## 7. Visual Examples

Consider Figure 4 where we present four specific examples showing the resulting visual relationships recognized by RONNIE from four specific images in VG-Min100. In each example, the orange arrow corresponds to a relationship that was not present in the training set (but was present in the auxiliary text). As seen in the top left image, our model fails to recognize the correct relationship label (`reading`) and predicted `reads`. While such errors occur in both seen and unseen cases, the predicted relationship is very often semantically related.

## 8. Ablation Study

**Importance of RONNIE's building blocks** We test each of RONNIE's building blocks on VG-200 in the SG-CLs setting to demonstrate the importance of each block. Refer to Table 4 for the results. From the table we conclude that both the object attention blocks and the relationship attention blocks are vital for the SGG task.

**Importance of auxiliary text** To demonstrate RONNIE's and RONNIE$_{gut}$'s superiority when recognizing unseen classes, we use RONNIE$_0$, with the only difference between the two models being the auxiliary text utilization. In Table 1 we present our results on VG-Min100. RONNIE demonstrates superiority over RONNIE$_0$ on VG-Min100. When evaluated on all relationships, RONNIE outperforms RONNIE$_0$ in top-5, top-10 accuracy, R@50 and R@100 by at least 3.75%, 2.6%, 8% and 7% respectively. When evaluated on rare relationships (occurring less than 1,024 times in the dataset), the impact is even greater as the performance margins grows and RONNIE$_{gut}$ outperformed all other models by at least 6.5%.

## 9. Concluding Remarks

We presented RONNIE, a novel approach that leverages text for recognition of visual relationships and composing scene graphs. One difficulty characteristic of visual relationship recognition systems is their inability to scale-up to large vocabularies. We demonstrated how leveraging auxiliary linguistic knowledge enables recognition of unseen classes and facilitates large-scale VRD. We also demonstrated how auxiliary knowledge induces recognition of classes within the long-tail of the visual relationship distribution. A surprising result was the performance of the Project Gutenberg-based model that was comparable to the image-based model and even outperformed the image-based model on rare relationships. Accordingly, we can conclude that leveraging auxiliary text could be beneficial even if the text is from a different domain, such as books. An interesting enhancement of our approach would be to try and learn the auxiliary linguistic information. One way to tackle this idea would be to follow Radosavovic et al. (2018) and utilize unlabeled data on a trained system such as RONNIE to obtain pseudo labels from which we could assemble an auxiliary text. We leave this for future work.